# A synthetic biology approach for the design of genetic algorithms with bacterial agents

A. Gargantilla Becerra[a]   M. Gutiérrez[b]   R. Lahoz-Beltra[c]*

a. Department of Systems Biology, Centro Nacional de Biotecnología (CSIC), 28049 Madrid (Spain)

b. Escuela de Informática y Telecomunicaciones, Universidad Diego Portales, Santiago (Chile)

c. Department of Biodiversity, Ecology and Evolution (Biomathematics), Faculty of Biological Sciences.  c/ Jose Antonio Novais 2, Complutense University of Madrid, 28040 Madrid (Spain)

\*   Correspondence: lahozraf@ucm.es

Abstract—Bacteria have been a source of inspiration for the design of evolutionary algorithms. At the beginning of the 20th century synthetic biology was born, a discipline whose goal is the design of biological systems that do not exist in nature, for example, programmable synthetic bacteria. In this paper, we introduce as a novelty the designing of evolutionary algorithms where all the steps are conducted by synthetic bacteria. To this end, we designed a genetic algorithm, which we have named BAGA, illustrating its utility solving simple instances of optimization problems such as function optimization, 0/1 knapsack problem, Hamiltonian path problem. The results obtained open the possibility of conceiving evolutionary algorithms inspired by principles, mechanisms and genetic circuits from synthetic biology. In summary, we can conclude that synthetic biology is a source of inspiration either for the design of evolutionary algorithms or for some of their steps, as shown by the results obtained in our simulation experiments.

# 1. Introduction

Evolutionary computing is a discipline of computer science, which aims to design algorithms inspired by the Darwinian evolution of organisms. However, many evolutionary algorithms exploit another principle and, in particular, the massive and parallel search for a good solution. Of all the organisms that exist in nature, bacteria are agents showing both principles as well as a remarkable evolutionary potential, standing out mainly for their fast evolution and high adaptability. Furthermore, bacteria display a wide repertoire of genetic mechanisms that have been a major source of inspiration in evolutionary computing.

Bacteria-inspired evolutionary algorithms arise from the need to resolve some of the distinctive setbacks of optimization methods. At the end of the 90s of the last century, [1] introduced a Bacterial Evolutionary Algorithm with new genetic operators for the simulation of gene transfer and mutation, and [2] Microbial Genetic Algorithm, which includes a recombination operator inspired by bacterial conjugation. More recently, genetic algorithms were designed to solve specific optimization problems with gene transfer operators and non-standard versions of genetic mutation operators, e.g. inverse mutation and pairwise interchange mutation [3]. The above algorithms illustrate some examples where bacteria provide a source of inspiration for new genetic operators. In fact, bacteria have the ability to transfer genes between individuals of the same generation, which is known as horizontal gene transfer. An example of horizontal gene transfer is the bacterial conjugation mechanism. For instance, [4] introduced a bacterial conjugation operator showing its usefulness in the design of an AM radio receiver. Afterwards other bacterial conjugation operators were introduced [5], exploring the possibility of incorporating physiological behaviors of bacteria into an evolutionary algorithm. For example [6] incorporates the chemotactic behavior of *E. coli* bacteria which is one of the main steps in the Bacterial Foraging Optimization Algorithm [7], which is one of the most distinctive bacteria-inspired algorithms. New versions of evolutionary algorithms based on bacteria are designed by their hybridization with other techniques, e.g. the Bacterial Memetic Algorithm [8] includes local search methods, particularly the Levenberg-Marquardt method.

On the other hand, advances in synthetic biology have led to a new perspective on bacteria, considering these microorganisms as organisms in which it is possible to program specific tasks. Today, these bacteria, known as synthetic bacteria, are programmable microorganisms with applications in biomedicine, pharmacology, bioremediation, bioenergy, etc. [9, 10, 11]. The most common approach, known as 'top-down' strategy, involves the use of a microorganism, for example a bacterium, into which we insert elements from the outside. At present, synthetic biology assumes that bacteria are computational agents or machines [12]. Thus, a bacterium receives inputs of a chemical or physical nature that are processed through 'biochemical hardware', which in turn is regulated by algorithms controlling gene expression. In this context, the evolution of a bacterial population towards an optimal state is accomplished through a protocol known as programmed evolution. At present, programmed evolution techniques make it possible to carry out *in vitro* evolution experiments using for this purpose those ingredients that are distinctive of Darwinian natural selection, i.e. bacterial population variability and artificial selection based on bacterial fitness. Fitness can adopt multiple different forms such as life/death of a bacterium, growth rate, or even protein of interest concentration. In 2012 [13, 14, 15] a programming language called Gro was introduced, allowing the modelling, specification and simulation *in silico* of the behavior of synthetic bacteria growing in a colony. At present, agent-based modelling [16, 17] is one of the common techniques in the prototyping and design of simulation experiments in synthetic biology.

In this work, we explore the application of synthetic bacteria colonies in evolutionary computation, by designing evolutionary algorithms conducted by bacterial agents. Bacterial agents were programmed using the Gro language, introducing a genetic algorithm in which all stages of the algorithm are performed by a colony of synthetic bacteria. The algorithm is termed BAGA, an acronym for Bacterial Agent Genetic Algorithm, simulating the evolution of a colony of bacterial agents. The utility of BAGA algorithm is evaluated with classical optimization problems, showing the practical utility of three important features present in both synthetic bacteria and their natural counterparts. First, bacteria are reproduced by binary excision, i.e. the colony grows exponentially or according to a Malthusian model [18]. Consequently, a large number of individuals ensures a massive exploration of the solutions space. Secondly, and since individuals are autonomous agents, the exploration of the environment is in parallel. Thus, combining these two features bacteria can find the solution to a given optimization problem [19] by checking all possible solutions. Thirdly, the pressure of natural selection acts on the bacterial colony changing the value of the Malthusian parameter of the equation ruling the exponential growth of the bacterial colony. As a result, the bacteria that evolve are those that increase their growth rates [20]. In line with these principles, BAGA can be tuned to seek an optimal solution by performing a massive search, by natural selection or by both principles simultaneously.

In Section 2 of this paper, we present the methodology and simulation experiments description solving several optimization problems, and in Section 3 we present the whole results of the computer simulation experiments. Finally, Section 4, discusses the possible impact of this work together with the general limitations.

## 2. Methods

In the present work, bacteria are agents programmed in a cellular programming language, in particular Gro 4.0 [13]. Using Gro language, we coded the scripts of both the BAGA algorithm and in the interactive evolutionary algorithm the screening procedures. Bacterial colony executes a program written in this language parallel to each bacterium, allowing the emergence of collective behaviors and different biological phenomena, e.g. cell division, chemotaxis, signal diffusion, etc. [13]. An extended version of Gro has recently been published [15] that allows plasmids recombination. Although the former feature is not used in our simulation experiments, we have used this version because the execution of the scripts is slightly faster than the original version of Gro.

### *2.1. A bacterial agent genetic algorithm*

BAGA, an acronym for <u>B</u>acterial <u>A</u>gent <u>G</u>enetic <u>A</u>lgorithm, is a genetic algorithm inspired in the evolution *in silico* of a colony of synthetic bacteria or bacterial agents. Bacteria efficiently explores the evolutionary surface looking for the optimal solutions through two features present in bacterial colonies. Firstly, the bacteria reproduce by bipartition growing the population according to an exponential growth curve. This feature allows the bacterial colony to search for the optimum or quasi-optimal solutions massively and in parallel. Secondly, bacteria selection takes place according to their fitness, because of the influence that fitness has on the growth rate. For example in bacteria such as *E. coli* the higher the fitness, the greater the growth rate value will be. Thus, in nature bacterial evolution by Darwinian natural selection is the result of an adjustment of the expression level of proteins (e.g. enzyme activity, i.e. proteins that act as biological catalysts for chemical reactions). Such adjustment has an effect on the bacterial growth rate [20] or Malthusian parameter $k$ in the equation ruling population growth:

$$y(t) = y_0 e^{kt} \qquad (1)$$

where $y(t)$ is the number of bacteria at time $t$, and $y_0$ the initial number of bacteria. The goodness of the solution found by a bacterium is detected and evaluated with a reporter circuit. This kind of genetic circuit is included into the plasmid of synthetic bacteria, detecting the best solutions according to the fluorescence emitted by the bacteria. Hence, the different degree of fluorescence observed in a colony of bacteria will show the fitness of each bacterium and therefore in some way the colony evolutionary landscape.

### 2.1.1. Coding

Similar to genetic algorithms and other evolutionary algorithms, solutions are encoded in a string of characters. Since the agents simulate bacteria, the string represents a plasmid. Plasmids are circular strands of DNA that are different from chromosomal DNA. Present in bacteria, plasmids carry genes that give adaptive advantage to bacteria; when a bacterium reproduces by bacterial division each daughter bacterium receives a copy of the plasmid. Manipulation of plasmids allows designing customized gene circuits, being this one of the main goals of synthetic biology.

In this paper, a plasmid as a list in Gro being either a binary encoding, permutation encoding (numbers representing a sequence) or a string of values (e.g. characters, real numbers, words etc.). For example, a binary plasmid and a real number plasmid are defined as:

$$plasmid := \{1,0,1,0,1,1,0,0\} \quad plasmid := \{A,B,C,D,E\}$$

In some optimization problems, it may be necessary the concatenation of lists in Gro language:

$$plasmid := \{0,1,4\}\#\{2\}\#\{5,1,6\}\#\{2\}\#\{7,3,8\}\#\{2\}\#\{5,3,6\}\#\{2\}$$

### 2.1.2. Standard BAGA algorithm

In this section, we show the main steps of the standard version of BAGA algorithm (Fig. 1):

*Step* 1. *Initial bacterium setup.*- In $t=0$ a single bacterium is set by assigning the initial values to the following parameters:

(1) Set the initial value of the growth rate: $k_0$.
(2) Initialize to zero the values of the GFP (green fluorescent protein) and IPTG (isopropyl β-D-1-thiogalactopyranoside) concentrations: $gfp=0$, $iptg=0$.
(3) Set to zero the initial fitness value: $z=0$.
(4) Given an optimization problem, we will encode the potential solutions into a plasmid $plasmid=\{x_1,x_2,\ldots,x_l\}$. In the simplest case, and if we use a binary sequence, e.g. $plasmid=\{0,1,\ldots,1\}$ then in the initial bacterial agent the plasmid is initialized with a list of zero values, i.e. $plasmid=\{0, 0, \ldots,0\}$.

*Step* 2. *Bacterial division.*- Bacteria reproduce by cell division yielding two bacterial cells. One of them referred to as the mother cell, the other daughter cell, although in fact the two are indistinguishable.

*Step* 3. *Mutation.*- After cell division, one of the bacterial cells retains the plasmid genes state while the other daughter bacterium undergoes point mutations. It is also possible for both cells to undergo mutation. In the algorithm, this is an option to set in the Gro script, since it affects the evolution of the bacterial colony. Mutation operator is a flip-bit operator, setting a $p_m$ value for the mutation rate.

*Step* 4. *Fitness evaluation.*- Once the bacterial division and mutation take place, the fitness value is calculated (Fig. 2). Inspired by synthetic biology BAGA assumes that bacteria include in their plasmid a hypothetical operon (i.e. a group of genes under the control of a region of the DNA called a promoter) as well as a molecule playing a role of operon activator. When activator enters into the interior of the bacterium, it activates the operon being the activator concentration the output of a particular optimization problem. For example, let us consider a problem consisting in finding the maximum of a function $y=f(x)$, then the activator concentration is given by $y$. In this example, the value of *x* results from decoding the plasmid sequence $\{x_1, x_2, ..., x_l\}$. In terms of synthetic biology, the algorithm assumes as activator the sugar emulator IPTG (isopropyl β-D-1-thiogalactopyranoside), which activates Lac operon (i.e. the operon lactose). For instance, in the above optimization example, the concentration of IPTG is the value of *y*, in the 0/1 knapsack problem IPTG concentration is given by $\sum_{i=1}^{l} v_i \cdot x_i$ where $x_i$ represents whether a certain item *i* has been (1) or not (0) included in the knapsack, being $v_i$ its value, etc. In short, in the algorithm IPTG molecules act as an activator of a 'Z operon'. The Z operon calculates the fitness of a given bacterium from the expression of a Z gene that synthesizes a *z* protein. The concentration of *z* protein, is the bacterial fitness value, which is given by function *z=f*(IPTG). The convenience of using a particular function will depend on the optimization problem. For instance, linear function, Hill function or any other function may be appropriate on a case-by-case basis (Table I). In the case of the Hill function and this applies to any other chosen function, its parameters (i.e. *v*, *k* and *n*) are set empirically, normalizing with *z=f*(IPTG) function the value of fitness *z* between 0 and 1.

| Linear function | $z = \dfrac{k \cdot [iptg]}{s}$ |
|---|---|
| Hill function | $z = \dfrac{v \cdot [iptg]^n}{k^n + [iptg]^n}$ |

Table 1. Fitness functions *z=f*(IPTG) for BAGA algorithm.

*Step* 5. *Screening method*.- As the bacterial colony grows, bacterial agents will report fitness *z*. Solution goodness coded in the bacterial plasmid, will be expressed through synthesis of fluorescent proteins in bacteria. In the algorithm, we assume that concentration of green fluorescent protein, i.e. GFP, is synthesized in function of the fitness value *z* (Fig. 2). Thus, since the Z operon activates in turn an operon reporter that expresses GFP protein, the higher the fitness value, the better the solution and therefore the greater the fluorescence that a bacterium emits:

$$gfp = m \cdot z \quad (2)$$

with *m* being the proportionality constant.

*Step* 6. *Selection*.- Bacterial selection is based on growth rate. Z operon not only influences the GFP fluorescence emitted by bacteria, but also affects the Malthusian parameter or bacterial growth rate (*k*). Updating of bacterial growth rate is given by the following expression:

$$k = k_0 + z\frac{\alpha}{\beta} \quad (3)$$

being $k_0$ the initial growth rate at *t*=0 and $\alpha$, $\beta$ two parameters for setting *k* parameter in simulation experiments. Notice how in the particular case of $\alpha$ =0 and $\beta$ =1 then BAGA algorithm performs optimization in the absence of selection and therefore by the parallel and massive search.

### 2.1.3. BAGA algorithm with penalty: solving the 0/1 knapsack problem

In this paper, we also designed an improved version of BAGA. This version of the algorithm is useful when an optimization problem requires a penalty, e.g. the knapsack problem (Fig. 3). Next, we will introduce the improved BAGA algorithm by adopting as an example the 0/1 knapsack problem. Let $w$ be the weight of the knapsack, i.e. $\sum_{i=1}^{l} w_i \cdot x_i$ after entering the selected objects $x_i$ with weights $w_i$, and $W$ the maximum tolerated weight. The improved BAGA algorithm includes penalty and shares the same steps as the standard BAGA, with the exception of the following two features. Firstly, the improved BAGA instead of simulating the concentration of IPTG simulates $v_0$, thus the velocity at which IPTG enters into the bacterium. In the knapsack problem, the input velocity of IPTG represents de profit calculated by $\sum_{i=1}^{l} v_i \cdot x_i$. Secondly, the fitness calculation (*Step* 4) is conducted depending on whether or not the constraint imposed on the optimization problem is met:

*Step* 4. *Fitness evaluation in solutions that do not satisfy the restriction ($w>W$).-* We will consider those solutions where the weight of the knapsack exceeds the maximum permissible weight. In such cases, the error $w-W$ value is simulated as the concentration of an inhibitor, i.e. a phosphotransferase IIA protein [21], which inhibits the entry of IPTG into bacteria (Fig. 3):

$$Inhibitor = w\text{-}W \quad (4)$$

The velocity at which the IPTG enters into the bacterium is given by the following kinetic expression:

$$v_0 = v \frac{[iptg]}{[iptg] + k\left(1 + \frac{Inhibitor}{k_2}\right)} \quad (5)$$

calculating the value of $k$ as the ratio between the sum of the values of the objects and the length $l$ of the plasmid:

$$k = \frac{\sum_{i=1}^{l} v_i}{l} \quad (6)$$

When we use in the model the Hill function, then we get the following fitness function:

$$z = \frac{v \cdot v_0^n}{k^n + v_0^n} \quad (7)$$

Notice how in the above expressions $k$ represents the Michaelis' constant $K_m$ and $v$ the maximum velocity $V_{max}$.

*Step* 4. *Fitness evaluation in solutions that satisfy the restriction ($w \leq W$).-* When a solution satisfies the restriction, i.e. the weight of the knapsack is below the maximum allowed value $W$ then the algorithm assumes that inhibitor is not present. In this case, the velocity of IPTG intake into the bacterium is given by the following expression:

$$v_0 = v \frac{[iptg]}{k + [iptg]} \quad (8)$$

calculating $k$ with expression (6) and fitness as we did before with the Hill function:

$$z = \frac{v \cdot v_0^n}{k^n + v_0^n} \quad (9)$$

In summary, for the example of the 0/1 knapsack problem, the fitness in the improved version of BAGA is given by the following expression:

$$z = \frac{v \cdot v_0^n}{k^n + v_0^n} \begin{cases} if\ w > W, & v_0 = v \dfrac{[iptg]}{[iptg] + k\left(1 + \dfrac{Inhibitor}{k_2}\right)} \\ if\ w \leq W, & v_0 = v \dfrac{[iptg]}{k + [iptg]} \end{cases} \quad (10)$$

### *2.1.4. A customized BAGA algorithm to solve the Hamiltonian path problem*

One of the optimization problems we applied the BAGA algorithm was the replication *in silico* of the synthetic biology experiment carried out with real bacteria [22, 23]. In this experiment, a colony of bacteria is programmed as a computer, solving an elementary instance of the Hamiltonian path problem. The *in vivo* experiment illustrates how to encode a problem in the DNA chain of a bacterium. In bacteria, gene components must be in the proper orientation (which is known as 5' → 3'). The components of a bacterial gene [24, 25] are promoter, ribosome (i.e. a complex particle where protein synthesis takes place) binding site (rbs), the structural gene (i.e. the actual gene responsible for the synthesis of a protein) and the end 3' or transcriptional terminator (TT) (i.e. a point at which ends the expression of a gene).

We successfully reproduced the general methodology adopted in the *in vivo* experiments, solving the three-node Hamiltonian path problem (Fig. 4). In the real experiments, their authors [22] coded the problem into a plasmid by splitting in two segments the genes that synthesize fluorescence-emitting proteins. In particular, the gene of GFP was split into two sections (Fig. 5), one referred to as right or 3' GFP (GFPr) and the other as left or 5' GFP (GFPl). The other gene was involved in the synthesis of the red fluorescent protein (RFP), which was also broken down into two parts: 3' RFP or RFPr and 5' RFP or RFPl. Following, [22] found a procedure by which it would be possible to permute gene sections, conducting a crossover of the gene segments inside the bacterium, a method known as Hin-hixC recombinase. In this paper, we designed a customized operator (Fig. 6) simulating this recombination mechanism. The application of this operator in the bacterial colony conducted to different random sequences of gene segments (Fig. 5), and therefore of the graph edges. We will refer to the joining points of the segments as HixC. For example, in the AC path when gene for RFP protein is expressed, then the bacterium emits red fluorescence. When a bacterium found the only Hamiltonian pathway in the graph, i.e. the AB pathway, then both GFP and RFP genes are expressed, synthesizing the green and red fluorescent proteins. In consequence, the bacteria with the optimal solution will emit yellow fluorescence. Simulation experiments with BAGA algorithm were conducted as follows:

*Step* 1. *Initial bacterium setup.*- Bacterial plasmid was initialized by concatenating lists representing the gene segments. The plasmid obtained, named as plasmid Hin/hix, encodes the graph into the bacterial plasmid according to the code shown in Table II:

*plasmid*= {0,1,4}#{2}#{8,8,8}#{2}#{8,8,8}#{2}#{8,8,8}#{2}

In accordance with the defined code the paths A, B and C of the graph were A= {5,1,6}, B= {7,3,8} and C= {5,3,6}.

| Gene component | Code |
|---|---|
| promoter | 0 |
| rbs | 1 |
| HixC | 2 |
| TT | 3 |
| 5' RFP | 4 |
| 3' RFP | 5 |
| 5' GFP | 6 |
| 3' GFP | 7 |
| null | 8 |

Table 2. Hin/hix plasmid code

*Step* 2. *Hin-hixC recombinase operator.-* Once the bacterial division has concluded, we apply the genetic operator simulating the Hin-hixC recombinase. The algorithm begins by randomly choosing two $x$, $y$ positions in the plasmid that will undergo the exchange, i.e. $(x,y) \rightarrow (y,x)$. Notice how this DNA exchange mechanism occurs within the plasmid of a bacterium, not between the DNA of two different plasmids. The value of $x$ is obtained randomly, whereas the value of $y$ is obtained from a random number $u \in [0, 1]$, which is compared with a $p$ probability value ($p$=0.5). The exchange between DNA segments is then performed ($u<p$) according to chart depicted in Fig. 6. In the simulation experiments, the Hin-hixC recombinase operator is applied with a probability $p_{Hix}$.

*Step* 3. *Expression of reporter genes.-* After obtaining the new configurations of the plasmid DNA strands (Fig. 5), the algorithm proceeds expressing the genes of the fluorescent proteins. Note how the presence of 4}#{2}#{5 sequence stands for the binding of the two halves of the gene encoding for RFP protein, emitting the bacterium a red fluorescence. Similarly, the binding of the two halves of the GFP gene is revealed by the pattern 6}#{2}#{7, emitting the bacterium a green fluorescence. However, when the respective halves of both genes are properly bound together, then a simultaneous emission of red and green fluorescence will result in a yellow light emission in the bacterium. Only bacteria that emit yellow fluorescence have found the three-node Hamiltonian pathway solution.

### 2.1.5. Simulation experiments

In order to test the potential usefulness of the algorithms described above we conducted the following simulation experiments. A first experiment was an optimization problem consisting of finding the maximum value of the function:

$$y = \frac{x-5}{2 + \text{Sin } x} \quad , \quad 0 \leq x \leq 16 \quad (11)$$

First, at time $t$=0 (*Step* 1) in the initial bacterium a binary a *plasmid* = {$x_0$, $x_1$, $x_2$, $x_3$} is obtained randomly. Following, the bacterium undergoes a bipartition (*Step* 2) and mutation with $p_m$=0.3 (*Step* 3) in the plasmid of the daughter bacterium, the other cell is the parental bacterium. Next, the plasmid is expressed, translating the binary sequence $x_0 x_1 x_2 x_3$ to its equivalent decimal number, resulting the $x$ value of the function (11). The concentration of IPTG or $y$ is given by $y=f(x)$, obtaining the fitness (*Step* 4) with a linear function (Table I) where $k$=10 and $s$=60. Afterwards, GFP value will depend on the fitness $z$ value (*Step* 5), with $m$=150. Finally, the Malthusian growth rate $k$ is updated (*Step* 6), being $k_0$=0.03, $\alpha$ =0.8 and $\beta$ =10.

In a second experiment, the goal was to find the minimum value of the booth function:

$$y = (x_1 + 2x_2 - 7)^2 + (2x_1 + x_2 - 5)^2 \quad , \quad 0 \le x_1 \le 8, 0 \le x_2 \le 8 \quad (12)$$

We defined a longer plasmid, *plasmid* = {$x_0, x_1, x_2, x_3, x_4, x_5$}, since (12) is a two variables function, encoding the first three bits for the variable $x_1$ and the remaining three bits for $x_2$. The experiment was conducted with the following parameter values: $p_m$=0.5, $k$=10, $s$=7000, $m$=150, $k_0$=0.03, $\alpha$ =0.8 and $\beta$ =1.0.

In the optimization experiments described above, we evaluated four variations of the standard BAGA algorithm. Firstly, in expression (3) when $\alpha$ >0 and $\beta$ >0 then we set the conditions of the default version searching for the optimal solution via selection (S) and parallelism (P). Secondly, by setting $\alpha$ =0 and $\beta$ >0 in (3) selection is removed conducting only a parallel (P) search for the optimal solution. Finally, in both cases it is possible to include a procedure that we termed as eugenic (E) procedure. In such a case, an additional step is included in the algorithm by which are removed those bacterial agents where GFP reporter protein levels are below to a certain threshold $\theta$ value. Thus, the algorithm kills bacteria without good solutions (*gfp* $\le \theta$). In this paper, we will run the algorithm in one of these four scenarios, which we will refer to as: (1) SP when the search for the optimum is via selection and parallelism, and (2) SPE when in (1) the eugenic rule is included. Other options are (3) P if the search for the optimum is in the absence of selection, with only parallelism, and (4) PE when in (3) the eugenic rule is included.

Following, we evaluated the performance of the improved version of BAGA comparing the obtained results with the standard BAGA algorithm. In this case, both experiments were just run according to SP protocol and solving an instance of the 0/1 knapsack problem (Table III), with *W*=100. In the standard BAGA experiments, we set $p_m$=0.3, $m$=150, $k_0$=0.03, $\alpha$ =2.0 and $\beta$ =10.0, obtaining the bacterial fitness with a Hill function (Table I) where *v*=1.0, *k*=27 and *n*=6.0. Next, we conducted the experiments with the improved version of BAGA. The instance and general parameter values were similar to those used with the standard BAGA, with the exception of the values of the parameters related to the fitness value calculation. Thus, parameters of the Hill function were *v*=1.0, *k* was obtained according to (6) with *n*=3.0 and $k_2$=0.02.

| Profit | 50 | 55 | 35 |
|---|---|---|---|
| Weight | 65 | 45 | 55 |

Table 3.0/1 knapsack instance

Finally, simulation experiments solving the Hamiltonian path problem were conducted replicating the original experiment [22]. For this reason, there is no Darwinian natural selection and the Malthusian growth rate remains constant ($k_0$=0.03). Therefore, it is a P protocol experiment. Instead of mutation, the Hin-hixC recombinase operator was the source of variability of the bacterial colony, with $p_{Hix}$=0.3.

In all the experiments carried out, we recorded the optimal bacterial occurrence times. Using the program Statgraphics Centurion 18 version 18.1.12 we fit the experimental data to the following regression growth curve:

$$y(t) = e^{-a+bt} \quad (13)$$

An optimal bacterium contains a plasmid with a sequence encoding an optimal solution. These bacteria are detected by conducting a screening of the growing colony, thus a bacterium is optimal when GFP concentration (fluorescence emitted) is above a given threshold $\theta_{gfp}$. In the experiments with (11) and (12) functions we set $\theta_{gfp}$ =149, whereas in 0/1 knapsack problem a value $\theta_{gfp}$ = 145 was used. However, in the Hamiltonian path problem the detection of the optimal bacteria was conducted recording the times in which the bacteria that have arisen emit yellow fluorescence.

Scripts written in Gro language, which implement the algorithms (Sections 2.1.2, 2.1.3 and 2.1.4), and used to perform the simulation experiments described in section 2.1.5 can be downloaded from the repository [26].

## 3. Results

The results obtained with the different versions of BAGA, opens up the possibility of designing evolutionary algorithms based on synthetic bacteria *in silico*. Synthetic biology is a discipline still in development from which to draw inspiration for the design of evolutionary algorithms. In this paper, we conducted simulation experiments with colonies of bacterial agents, confronting the bacterial colony with different optimization problems. In order to specify whether the search for the optimal solution is exclusively in parallel or with the help of selection, simulation experiments were conducted by setting different values to the parameters of the BAGA algorithm. In Fig. 7, we show for the problem related with the optimization of function (11) a screenshot of the sequence of stages in the growth of the colony as well as the representative growth model $y(t)$ of the number of bacteria per time carrying optimal solutions (Fig. 8). Indeed, all variants of the standard BAGA algorithm lead to the formation of a population of optimal bacteria with a similar genome 1011, i.e. the value of *x* that maximizes the function (11). According to Table IV, the value of the estimated slope in the function modelling the population growth allows us to quantify the effects of selection in the search for the optimum, e.g. 0.071 in SP and 0.026 in P. In other words, for instance to obtain 246 bacteria with the optimal solution, it takes $t=196$ units of time for a SP search and $t=297$ for a P search. Therefore, when there is selection the bacteria find 101 units of time before the optimal solution. This means that as expected, and as BAGA is an evolutionary agent-driven algorithm, selection is a significant contributor in the parallel search for the optimal solution conducted by the bacterial agents. Interestingly, the inclusion in SP and P of such an extreme rule as the eugenic procedure leads to a delay in the emergence of bacteria carrying the optimal solution. However, after this delay, the number of optimal bacteria rises to the order of several thousand individuals with respect to the algorithms without this rule. When considering all four variants of the BAGA algorithm, the waiting time needed to reach a given number of bacteria carrying the optimal solution, e.g. for 246 bacteria, was SP ($t=196$) < P ($t=297$) < SPE ($t=511$) < PE ($t=675$).

Similar to the optimization problem described above in Fig. 9 we show the results of the experiments performed with standard BAGA in the search of the minimum of the booth function (12). In this simulation experiments the optimal bacteria carry the 001, 011 values that represent the $x_1$ and $x_2$ values which minimize the booth function. Although the selection is a factor that promotes the finding of optimal solutions, its effect is not as great in (12) as in the previous experiment with the function of one variable (11). One plausible explanation is that parallel search is the main factor driving bacteria during the search for the optimum. This is justified because regression function slopes are similar in the SP and P variants of the experiment (Table V), i.e. 0.036 and 0.035 respectively. The inclusion of a eugenic step in the BAGA algorithm further delays the presence of optimal bacteria (Fig. 9b) by adopting in SPE experiment a 'J' shape the growth function $y(t)$. Note how in this case the slope of the growth function exhibits a very high value (0.141) compared to the other population growth curves. Once again, in the four experimental variants of BAGA algorithm and given a number of bacteria, for example 88, the time required to observe such population size was SP ($t=236$) < P ($t=293$) < SPE ($t=870$) < PE ($t=1614$).

In the experiment carried out with the improved BAGA, thus the algorithm including punishment that was applied to 0/1 knapsack problem, the number $y(t)$ of bacterial agents carrying optimal solutions (Table VI) increases significantly (Fig. 10) with respect to the standard BAGA. Consequently, the replacement in the algorithm of the IPTG concentration by its velocity $v_0$ as well as the restriction ($w \leq W$) in (10) improve the rate at which bacterial carrying optimal solutions arise during the simulation experiment.

| Algorithm version | Regression curve [*] |
|---|---|
| SP | $y = e^{-8.492+0.071t}$ |
| SPE | $y = e^{-33.943+0.077t}$ |
| P | $y = e^{-2.291+0.026t}$ |
| PE | $y = e^{-10.251+0.023t}$ |

Table 4. Optimal bacteria growth models in the standard BAGA algorithm for the function (11) optimization problem. [*] The relationships between population size ($y$) and time ($t$) was statistically significant in all four models ($p$-value=0.0000)

| Algorithm version | Regression curve |
|---|---|
| SP | $y = e^{-4.236+0.036t}$ |
| SPE | $y = e^{-117.941+0.141t}$ |
| P | $y = e^{-5.727+0.035t}$ |
| PE | $y = e^{-25.022+0.018t}$ |

Table 5. Optimal bacteria growth models in the standard BAGA algorithm for the function (12) optimization problem. [*] The relationships between population size ($y$) and time ($t$) was statistically significant in all four models ($p$-value=0.0000)

| BAGA algorithm | Regression curve |
|---|---|
| Standard BAGA | $y = e^{-18.858+0.167t}$ |
| Improved BAGA | $y = e^{-7.327+0.098t}$ |

Table 6. Optimal bacteria growth models in the standard BAGA and improved BAGA algorithms for the 0/1 knapsack optimization problem. [*] The relationship between population size ($y$) and time ($t$) was statistically significant in all four models ($p$-value=0.0000)

Fig. 11 depicts the growth of the colony in the Hamiltonian path problem. The optimal growth curve of the number of bacteria finding the optimal pathway is shown in Fig. 12. In a particular experiment, after 262.65 time units, 318 bacterial agents found the optimal solution (Fig. 13). The results obtained in the *in silico* experiment replicate those obtained in the original *wet* experiment with real bacteria [22]. In the simulation experiment bacteria found the optimal solution in the absence of selection as occurs in its laboratory counterpart.

In summary, we can conclude that synthetic biology is a source of inspiration either for the design of evolutionary algorithms or for some of their steps, as shown by the results obtained in our simulation experiments. With the experiments conducted we have illustrated how using synthetic bacteria it is possible to solve some elementary optimization problems.

## 4. DISCUSSION

The results of the simulations are in line with those that could be expected from similar experiments conducted in the field of programmed evolution in both microbiology and synthetic biology. The different versions of the BAGA algorithm solve elementary instances of classic optimization problems in a similar way to a classical genetic algorithm, capturing the role played by the different components that are part of a synthetic biology experiment, which justifies the acceptance of the assumptions of the algorithm, as shown in the simulation experiments.

However, although the results are quite encouraging, there are still issues that need to be addressed in the future. In the experiments performed with the BAGA algorithm the only source of variability is the mutation. In particular, in the optimization experiments and 0/1 knapsack problem the mutation is similar to the operator mutation of a genetic algorithm. That is, given a certain probability of mutation, the operator randomly generates 0s and 1s that are substituted in the old gene value. Likewise, and similarly in the case of the Hamiltonian problem, the only source of variability in the BAGA algorithm is the mechanism of Hin-hixC recombination, as it occurs in its laboratory counterpart. Therefore, and in the future, the possibility of including in BAGA other sources of variability should be explored, and in particular we refer to the possibility of implementing bacterial conjugation, a form of recombination between neighboring bacteria. Bacteria are microorganisms that exhibit a variety of mechanisms for gene transfer between individuals, which are known as horizontal gene transfer [27]. Another interesting possibility not included in the present BAGA simulations will be the substitution of reporter proteins by other procedures for the detection of optimal solutions, for example the use of simulated antibiotics and the resistance of optimal bacteria to the presence of antibiotics.

One of the interesting aspects in the simulations is the the parallel search for solutions, thus a feature that is a natural consequence of the exponential growth of the bacterial population. The simulations suggest that although selection is an important factor in the evolution of the colony of synthetic bacteria, equally important is the parallelism. The mechanism of bacterial duplication leads to an exponential or Malthusian growth model in which bacteria of different generations explore the evolutionary space in search of the optimum. In the absence of the mechanism we have termed eugenic procedure, i.e. if we do not eliminate the bacteria carrying the worst solutions, then the entire colony in growth will collaborate in this search, and therefore in solving the optimization problem. Even if there are bacteria with optimal or near-optimal solutions that increase their growth rate, that is the value of the Malthusian parameter, the remaining agents will continue cooperating in the exploration of the solution space. Since in BAGA the fitness is expressed through the bacterial growth rate, then bacterial colony explores the evolutionary landscape in a fashion that best mimics natural selection. On the contrary, in a genetic algorithm all individuals are replaced generation after generation, with no overlap of generations. Therefore, the evolutionary dynamics of BAGA is different from the Wright-Fisher evolutionary dynamics of a genetic algorithm. Furthermore, the execution of any of BAGA implementations in the Gro cell programming language is a parallel run, since every daughter bacterium resulting from the division of a parent bacterium will be programmed with a copy of the script from the parent bacterium. Therefore, a script will be doubled itself as time goes by with the growth of the colony. For this reason, and given a certain problem of optimization the execution speed of the algorithm depends largely on the version of Gro used in the simulation as well as the computer's performance. A novel feature of BAGA is the replacement of the objective or fitness function characteristic of a genetic algorithm by a procedure in which fitness is calculated by a method inspired by synthetic biology. The possibility of obtaining a fitness value that depends on IPTG, a molecule that triggers the expression of any gene under the control of the Lac operon, corroborates how synthetic biology is a good source of inspiration in evolutionary computing. However, an important difference between our algorithm and a programmed evolutionary experiment in synthetic biology is that in the latter case, in an experiment with real bacteria in the laboratory, the selection of the bacteria is based on the idea that bacterial survival depends on gene expression. That is, the information of a gene is translated into a protein. In the case of an enzyme, i.e. a protein catalyzing a chemical reaction of the metabolism, then the expression of the gene is measured by the enzymatic activity. In contrast, in experiments with BAGA algorithm *in silico*, bacterial selection depends on the IPTG inducer and not on the gene expression, i.e. IPTG-induced enzyme activity. Nevertheless, BAGA shares some drawbacks with genetic algorithms. For instance, as it occurs in genetic algorithms, and regardless of whether the IPTG value expresses its concentration or velocity, we need to adjust the fitness value obtained either by scaling the values or by any other procedure. A similar situation occurs with $\alpha$ and $\beta$ parameters which allow the calibration of the effect of fitness on the bacterial growth rate. These values must be carefully adjusted to values that are within the physiological scales governing the dynamics of bacterial colony. In addition, the current version of BAGA assumes a linear relationship between bacterial growth rate and fitness. That is, in the expression (3) the increase in the growth rate is directly proportional ($\alpha$) to the fitness ($z$), which is scaled with the parameter $\beta$. Consequently, in the future it will be of interest to study other non-linear relationships between growth rate and fitness.

Another interesting question to study in the future is related with the fact that BAGA does not take into account whether the bacterial culture grows on a Petri dish, chemostat [28] or any other container [29]. The availability of space is one of the most relevant issues in the dynamics of a population of microorganisms, and from which could derive interesting issues related to the resolution of an optimization problem with bacterial agents. Related to the latter question, the potential emergence of spatial patterns from the application of BAGA to a bacterial colony can also be studied: strategies for harnessing the spatial patterns into improving the execution of the algorithm could be proposed, or conversely, producing spatial patterns from the algorithm as a secondary goal in establishing a blueprint for a primary goal such as biomaterial fabrication, biosensor implementation, or tissue engineering [30]. Other issues also to be studied are those related to the mutation that takes place after bacterial duplication. Each time a bacterium divides into two daughter bacteria, one of them retains the parental genes state while the other daughter bacterium undergoes point mutations. It is also possible for both cells to undergo mutation. In the algorithm, this is an option to decide and set, since it affects the evolution of the colony. Our goal has been to take the first steps in the design and development of a genetic algorithm, being the novelty of the algorithm that its main steps are inspired by the physiology, genetic mechanisms and dynamic behavior of synthetic bacteria. In the future we should address a study of all these issues, because then we would be able to design new versions of the current algorithm.

Although there are other programming languages [31] available in synthetic biology, both in BAGA and in the implementation of screening procedures, the programming of synthetic bacteria has been done with Gro language. We believe that this cellular programming language has a potential future as long as the language is maintained up to date, including new commands, libraries, etc. Likewise, and in our opinion, we believe that these kinds of simulation experiments will favor the opposite process, that is, when designing evolutionary algorithms not only do engineers look more like biologists, but biologists will become computer scientists [32]. Another possible effect of our work is in the field of synthetic biology. To date, in synthetic biology there is a predominance of the top-down approach, i.e. elements designed externally are added into a bacterium. However, one of the challenges of several scientific disciplines, e.g., synthetic biology, artificial life, bioinformatics, etc. is succeed with the opposite approach, i.e. a bottom-up approach [33]. The aim is the creation of an artificial cell or proto-cell from the gradual assembly of different elements, modules or systems. Therefore, our simulations could be a starting point about how to incorporate synthetic bacterial evolution in synthetic biology experiments [34], because it is likely that in a near future evolution will be one of the main ingredients of the bottom-up strategy. An exemplification of this scenario be applied to the automation of gene circuit designs. Electronic Design Automation (EDA) currently uses genetic algorithms for the placement of electronic components within a circuit, therefore, selecting the better circuits among many variants of possible ones. In 2016, a genetic design automation tool, Cello [35], was published. However, there is no mention to genetic algorithms in the implementation of Cello. An interesting direction could be to study the use of BAGA variants to automate the design process *in vitro/in vivo*.

Simulation experiments carried out with BAGA algorithm reproduce some of the main features of biological evolution. However, in the future, when BAGA-like algorithms are designed and applied to solving optimization problems *in vivo*, i.e. with real bacteria, some drawbacks will arise. In this sense, the experiments carried out by [20] as well as the study of microbial evolution in the laboratory [36] could provide interesting ideas for the *in vivo* version of BAGA. In fact, the possibility of implementing *in vivo* genetic algorithms has already been outlined. In 2005, [37] introduced an experimental protocol referred to as Cellular Evolutionary Computation by combining an *in vitro* phase where plasmids are recombined, and other phase *in vivo* where bacteria carrying the plasmids are selected according to their fitness. Nevertheless, their authors [37] are aware that the procedure is tedious, so with *in silico* simulations it is possible to reach similar results in a simpler way.

Finally, and as the main conclusion, we propose that synthetic biology opens up the possibility of developing new algorithms, genetic operators, and simulation protocols as a source of inspiration in the field of evolutionary computing. Furthermore, we believe that this kind of simulation experiments will someday help to make the leap into the design of evolutionary algorithms with real bacteria in e.g. a hardware system integrating a bioreactor with a computer.

## 5. REFERENCES

[1] M. F. Hatwágner and A. Horváth, "Maintaining genetic diversity in bacterial evolutionary algorithm," *Annales Univ. Sci. Budapest., Sect. Comp.* vol. 37, pp. 175-194, 2012.

[2] I. Harvey, "The microbial genetic algorithm," in *10th European Conference, ECAL 2009, Advances in Artificial Life. Darwin Meets von Neumann, Lecture Notes in Computer Science, Part II. LNCS 5778*, Berlin. Springer Verlag, G. Kampis, I. Karsai, and E. Szathmáry, Eds., Budapest, Hungary, 2011, pp. 126-133.

[3] C. Anandaraman, A. V. Madurai Sankar, and R. Natarajan, "A new evolutionary algorithm based on bacterial evolution and its application for scheduling a flexible manufacturing system," *Jurnal Teknik Industri*, vol. 14, pp. 1-12, 2012.

[4] C. Perales-Gravan and R. Lahoz-Beltra, "An AM radio receiver designed with a genetic algorithm based on a bacterial conjugation genetic operator," *IEEE Transactions on Evolutionary Computation*, vol. 12, pp.129 – 142, 2008.

[5] A. Mehrafsa, A. Sokhandan, G. Karimian, "A high performance genetic algorithm using bacterial conjugation operator (HPGA)", *Genet. Program. Evolvable Mach.*, vol. 14, pp. 395-427, 2013.

[6] P. K. Singh, S. K. Mondal, and M. Maiti, "A hybridized chemotactic genetic algorithm for optimization," *Journal of Computer Science Systems Biology*, vol. 9, pp. 45-50, 2016.

[7] K.M. Passino, "Biomimicry of bacterial foraging for distributed optimization and control," *IEEE Control Systems Magazine*, vol. 22, pp. 52-67, 2002.

[8] J. Botzheim, C. Cabrita, L.T. Kóczy, and A.E. Ruano, "Fuzzy rule extraction by bacterial memetic algorithms," *International Journal of Intelligent Systems*, vol. 24, pp. 312-339, 2009.

[9] E. Cameron, C.J. Bashor, and J.J. Collins, "A brief history of synthetic biology," *Nature Reviews Microbiology*, vol. 12, pp. 381-390, 2014.

[10] I.C. MacDonald and T.L. Deans, "Tools and applications in synthetic biology," *Advanced Drug Delivey Reviews*, vol. 105, pp. 20-34, 2016.

[11] M. Xie and M. Fussenegger, "Designing cell function: assembly of synthetic gene circuits for cell biology applications," *Nat. Rev. Mol. Cell. Biol.* 19**,** 507–525, 2018, doi.org/10.1038/s41580-018-0024-z.

[12] R. Lahoz-Beltra, J. Navarro, and P.C. Marijuan, "Bacterial computing: a form of natural computing and its applications," *Frontiers in Microbiology*, 5:101. DOI: 10.3389/fmicb.2014.00101, 2014.

[13] S.S. Jang, K.T. Oishi, R.G. Egbert, and E. Klavins, "Specification and simulation of synthetic multicelled behaviors," *ACS Synthetic Biology*, vol.1, pp. 365-374, 2012.

[14] K. Oishi and E. Klavins, "A framework for implementing finite state machines in gene regulatory networks", *ACS Synthetic Biology*, vol. 3, pp. 652-665, 2014.

[15] M. Gutiérrez, P. Gregorio-Godoy, G. Pérez Del Pulgar, L. Muñoz, S. Sáez, and A. Rodríguez-Patón, "A new improved and extended version of the multicell bacterial simulator gro," *ACS Synthetic Biology*, vol. 6, pp. 1496-1508, 2017.

[16] T.E. Gorochowski, "Agent-based modelling in synthetic biology," *Essays in Biochemistry*, vol. 60, pp. 325-336, 2016.

[17] A. Matyjaszkiewicz, G. Fiore, F. Annunziata, C.S. Grierson, N.J. Savery, L. Marucci, and M. di Bernardo, "BSim 2.0: An advanced agent-based cell simulator," *ACS Synthetic Biology*, vol. 6, pp. 1969-1972, 2017.

[18] R.J. Allen and B. Waclaw, "Bacterial growth: a statistical physicist's guide," *Rep Prog Phys.* 2019, vol.82:016601. doi:10.1088/1361-6633/aae546.

[19] V. Norris, A. Zemirline, P. Amar, et al., "Computing with bacterial constituents, cells and populations: from bioputing to bactoputing," *Theory Biosci.*, vol. 130, pp.211–228, 2011.

[20] E. Dekel and U. Alon, "Optimality and evolutionary tuning of the expression level of a protein," *Nature*, vol. 436, pp. 588-592, 2005.

[21] P. Hariharan, D. Balasubramaniam, A. Peterkofsky, and H.R. Kaback, "Thermodynamic mechanism for inhibition of lactose permease by the phosphotransferase protein IIAGlc," *PNAS*, vol. 112, pp. 2407-2412, 2015.

[22] J. Baumgardner, K. Acker, O. Adefuye, S.T. Crowley, W. DeLoache, J.O. Dickson, L. Heard, A.T. Martens, N. Morton, M. Ritter, A. Shoecraft, J. Treece, M. Unzicker, A. Valencia, M. Waters, A.M. Campbell, L.J. Heyer, J.L. Poet, and T.T. Eckdahl, "Solving a Hamiltonian path problem with a bacterial computer," *Journal of Biological Engineering*, vol. 3, pp. 1-11, 2009.

[23] J.L. Poet, A.M. Campbell, T.T. Eckdahl, and L.J. Heyer, "Bacterial computing," *XRDS*, vol. 17, pp. 10-15, 2010.

[24] L. Kari, R. Kitto, and G. Gloor, "A computer scientist's guide to molecular biology," *Soft Computing*, vol. 5, pp. 95-101, 2001.

[25] L. Hunter, "Molecular biology for computer scientists," in *Artificial Intelligence and Molecular Biology*, L. Hunter, Ed. AAAI Press/MIT Press, Menlo Park, CA, 1993, pp. 1-46.

[26] R. Lahoz-Beltra and A. Gargantilla Becerra, "A genetic algorithm designed with bacterial agents (BAGA)". figshare. Software. https://doi.org/10.6084/m9.figshare.13396544.v1, 2020.

[27] T.G. Villa, L. Feijoo-Siota, A. Sánchez-Pérez, J.R. Rama, C. Sieiro, "Horizontal gene transfer in bacteria, an overview of the mechanisms involved," in *Horizontal Gene Transfer*, T. Villa M. Viñas M, Eds. Springer, Cham, 2019, pp. 3-76.

[28] D. Gresham and J. Hong, "The functional basis of adaptive evolution in chemostats," *FEMS Microbiology Reviews*, vol. 39, pp. 2-16, 2015.

[29] D.E. Dykhuizen, "Experimental studies of natural selection in bacteria," *Annu. Rev. Ecol. Syst.*, vol. 21, pp.373-398, 1990.

[30] J. Santos-Moreno and Y. Schaerli. "Using synthetic biology to engineer spatial patterns." *Advanced Biosystems* 3(4), 2019: 1800280.

[31] P. Umesh, F. Naveen, C.U. Maheswara Rao, and A.S. Nair, "Programming language for synthetic biology," *Syst. Synth. Biol.*, vol. 4, pp. 265-269, 2010.

[32] A. Condon, H. Kirchner, D. Larivière, W. Marshall, V. Noireaux, T. Tlusty, and E. Fourmentin, "Will biologists become computer scientists?," *EMBO reports*, vol. 19:e46628, pp. 1-6, 2018.

[33] P. Schwille and K. Sundmacher, "Synthetic Biology: Life, Remixed," Available: https://www.mpg.de/8219292/synthetic_biology, Accessed on: Mar. 6, 2020.

[34] S.G. Peisajovich, "Evolutionary synthetic biology," *ACS Synthetic Biology*, vol. 1, pp. 199-210, 2012.

[35] A. A. K. Nielsen, B. S. Der, J. Shin, P. Vaidyanathan, V. Paralanov, E. A. Strychalski, D. Ross, D. Densmore, C. Voigt, "Genetic Circuit Design Automation," *Science* 2016, *352* (6281), aac7341, doi.org/10.1126/science.aac7341.

[36] M. Dragosits and D. Mattanovich, "Adaptive laboratory evolution – principles and applications for biotechnology," *Microb. Cell. Fact.* 12:64, 2013, doi.org/10.1186/1475-2859-12-64.

[37] K. Wakabayashi and M. Yamamura, "A design for cellular evolutionary computation by using bacteria," *Natural Computing*, vol. 4, pp.275-292, 2005.

**Álvaro Gargantilla Becerra** received the B.S. degree in biochemistry from the Córdoba University, Córdoba, Spain, in 2016 and Master's course degree on Environmental and Industrial Biotechnology from the Complutense University of Madrid, Madrid, in 2019. He is currently working at the National Center for Biotechnology as assistance researcher in the group of Systems Biotechnology.

He is interested in the study of evolutionary and ecological processes that occur within microbial populations. His recent research work has focused on the design of bioinspired algorithms for optimization problems and the development of metabolic reconstruction models of microorganisms. His long-term goals include the development of a methodology for predicting community-aggregated traits on microbiomes that are exposed to environmental changes and its application in microbiome engineering tools.

**Martín Eduardo Gutiérrez Pescarmona** received his B.Sc. in Industrial Engineering and M.Sc. in Computer Science from Pontificia Universidad Católica de Chile in 2005. In 2006, he was appointed as Lecturer at Universidad Diego Portales, Chile, where he taught all branches of Computer Science and specialized in AI techniques and meta-heuristic algorithms. Later, Martín completed his Ph.D. thesis from 2012 to 2017 at the AI Lab at Universidad Politécnica de Madrid,

Spain. During this period, his research was focused on Agent based Model bacterial colony simulators, synthetic gene circuits, bacterial conjugation, and spatio-temporal patterns in bacterial colonies. The main contribution of his thesis was extending the gro simulator. Dr. Gutiérrez briefly continued his work as a Postdoctoral researcher at the AI Lab of Universidad Politécnica de Madrid before returning to Universidad Diego Portales in 2018.

He is currently Assistant Professor at Escuela de Informática y Telecomunicaciones, where he continues his research on automation techniques applied to bacterial/cell colony simulators, interplay of Computer Science, AI and synthetic biology, spatio-temporal patterns and inter-cell communication based circuits.

**Rafael Lahoz-Beltra** received the B.S. degree in biological sciences and the Ph.D. degree in biostatistics from the Complutense University of Madrid, Madrid, Spain, in 1985 and 1989, respectively. He was a Fulbright Visiting Research Scholar in the Department of Systems Science, T. J. Watson School of Engineering, State University of New York at Binghamton, from 1989 to 1990, and continued his work under his Fulbright Scholarship from 1990 to 1992 in the Department of Anesthesiology, Health Sciences Center, University of Arizona, Tucson. He is an Associate Professor in the Department of Biodiversity, Ecology and Evolution (Biomathematics), Faculty of Biological Sciences, Complutense University of Madrid. He authored Bioinformática: Simulación, Vida Artificial e Inteligencia Artificial (Diaz de Santos, 2004), as well as the first two Spanish biographies of Alan Turing (Nivola, 2005; National Geographic, 2012). In 2008 worked for a trimester in the School of Computer Science, The University of Nottingham (UK) and in 2015 in the Dept. of Computing Science and Mathematics, University of Stirling (Scotland, UK). During the period 2015-2020, he worked as an expert evaluator of the EU's Horizon 2020 program.

He has performed research in cellular automata, the origin of life, synaptic regulation in biological and artificial neural networks, molecular automata in structural biology, cytoskeletal and microtubule participation in information processing. At present, his main research interests include evolutionary computation, cancer simulation models by hybridizing differential equations with AI techniques, embryo development modelling, quantum computing and the design of bioinspired algorithms. Dr. Lahoz-Beltra received in 2004 an award in the "2nd Contest of Spin-Off Companies proposed by Researchers from Madrid" by the project entitled Sinapgen Solutions S.L. He was awarded in 2012 by Parque Cientifico de Madrid for the spin-off project "Intelligent Affective Machines".

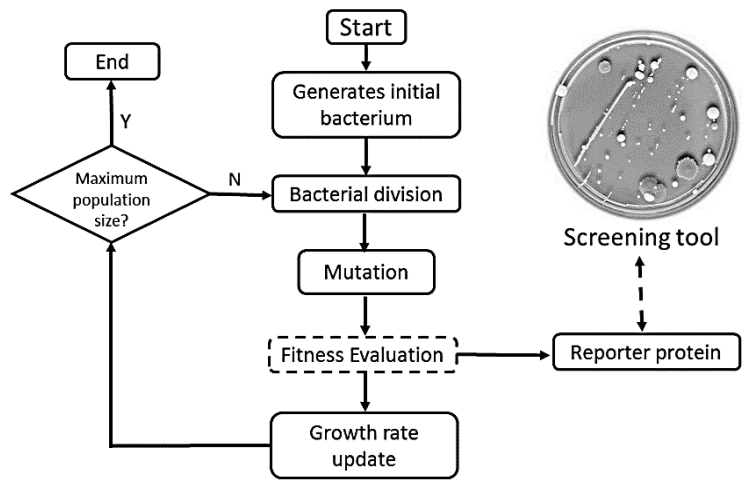

Figure 1. BAGA algorithm (for explanation see text).

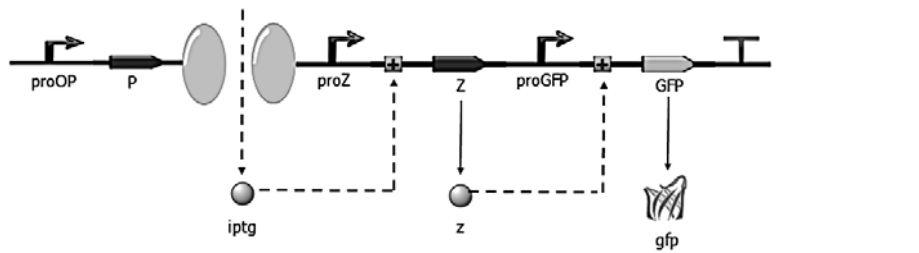

Figure 2. Fitness evaluation in standard BAGA algorithm. The P gene encodes a given optimization problem, e.g. finding the maximum of the function $y=f(x)$, being proOP the promoter. The concentration of IPTG entering the bacterium is the value of $y$, with x being the value/sequence coded in the plasmid. The IPTG value activates the 'operon Z' which calculates the fitness, value that is modeled by the synthesis of a protein $z$, i.e. $z = f(\text{IPTG})$. The fitness value is reported through the synthesis of a fluorescent protein, and in particular of GFP. The higher the fitness, the greater the GFP synthesis ( = promotor,  = gene,  = rbs,  = terminator,  = activator) .

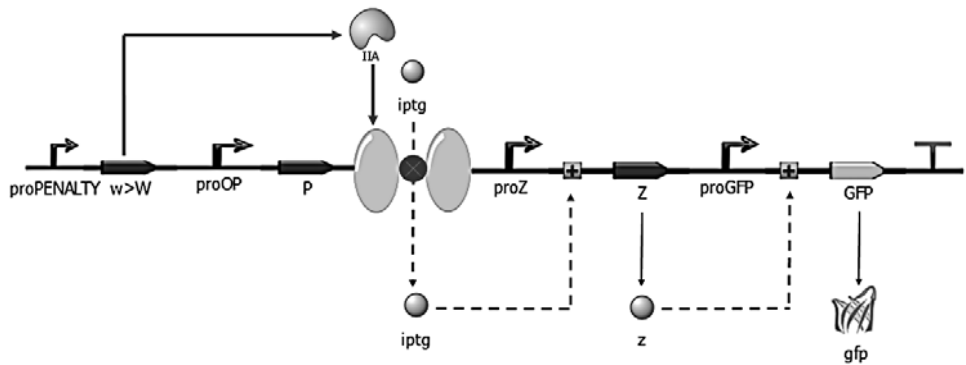

Figure 3. Fitness evaluation in BAGA algorithm with penalty. In the example the P gene encodes the 0/1 knapsack problem. The main stages are similar to those of the standard BAGA algorithm (Figure 2), except for the presence of an inhibitor, namely protein IIA. The protein is synthesized by a gene when w>W, being proPENALTY the promoter. If IIA is present then the entry of IPTG into the bacterium is blocked.

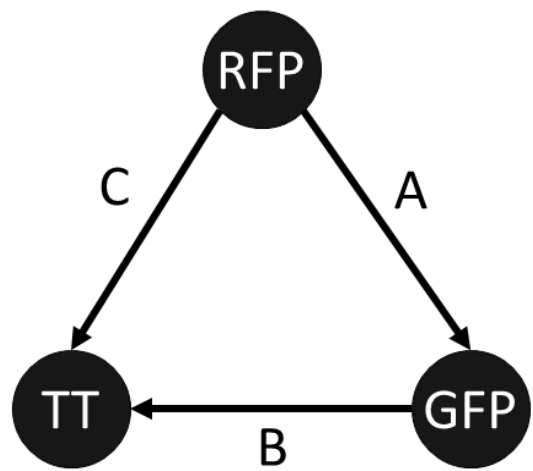

Figure 4. Three node Hamiltonian problem.

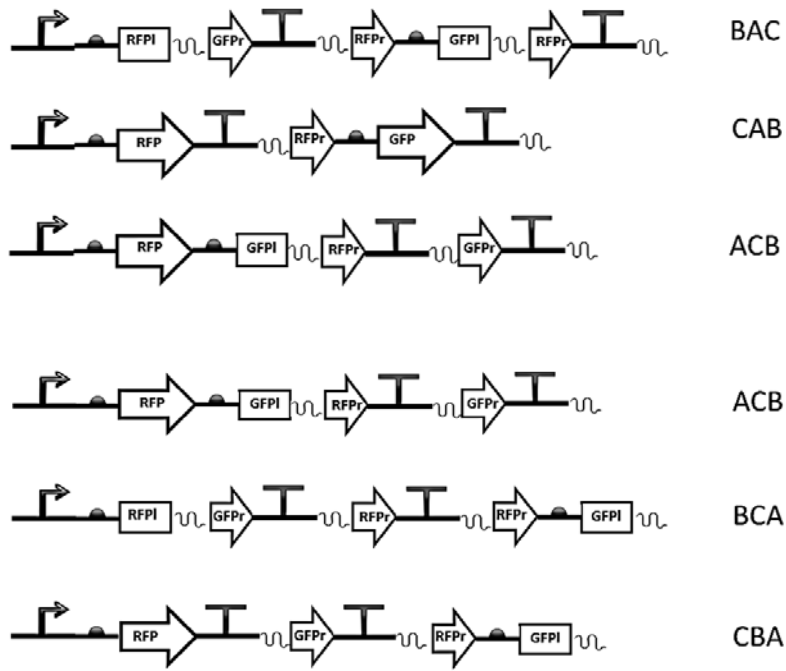

Figure 5. Solutions of Hamiltonian path problem obtained with BAGA algorithm (See Figure 2 for the meaning of the symbols, ∿ = HixC).

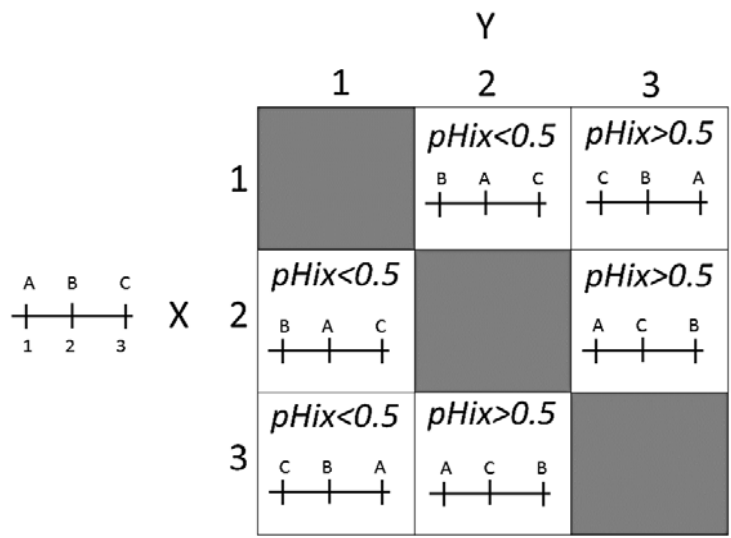

Figure 6. Hin-hixC recombinase operator (for explanation see text).

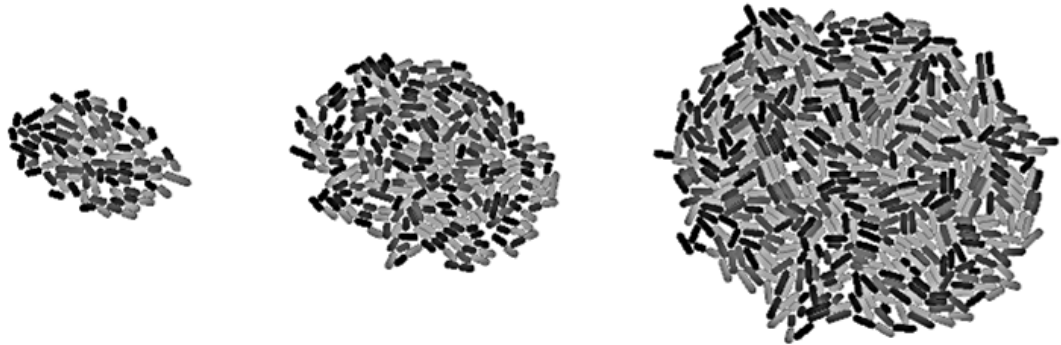

Figure 7. Growing colony for the optimization problem of the search of the maximum of the function (11). From left to right the colony is composed by 94 (*t*=667.31), 270 (*t*=714.35) and 524 (*t*=751.19) bacteria respectively. The different brightness of the bacteria depends on the concentration of GFP. The higher the fluorescence, the better the solution.

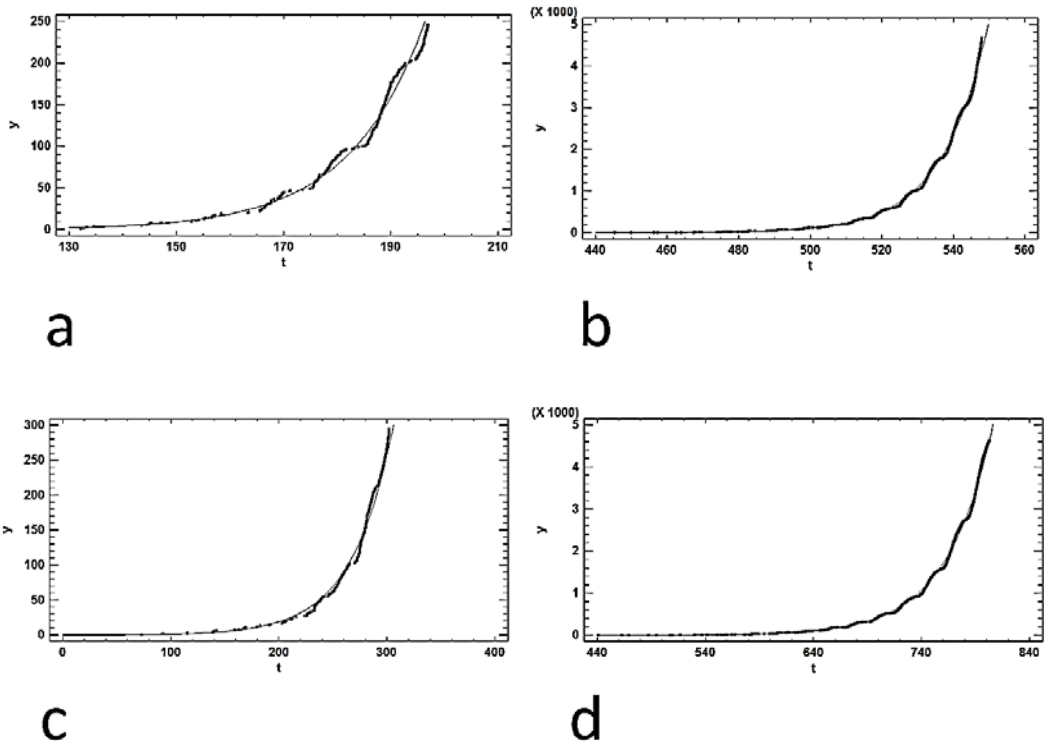

Figure 8. Number of bacteria bearing optimal solution (*y*) as a function of time (*t*) in the function optimization problem (11): (a) SP (b) SPE (c) P (d) PE.

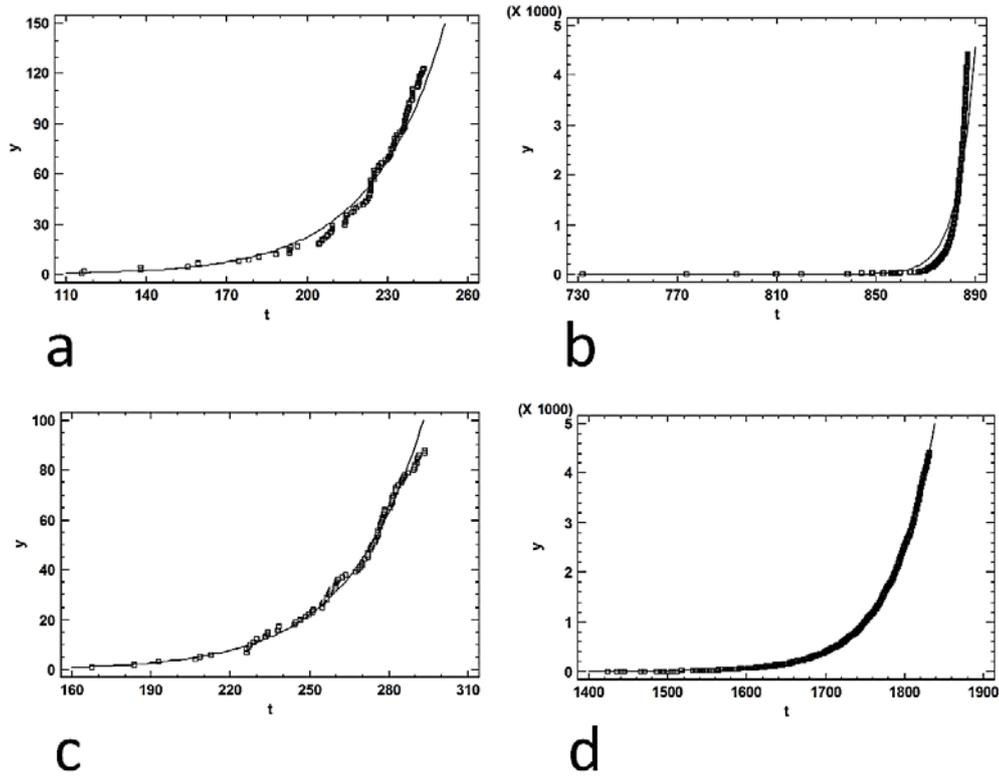

Figure 9. Number of bacteria bearing optimal solution (*y*) as a function of time (*t*) in the function optimization problem (12): (a) SP (b) SPE (c) P (d) PE.

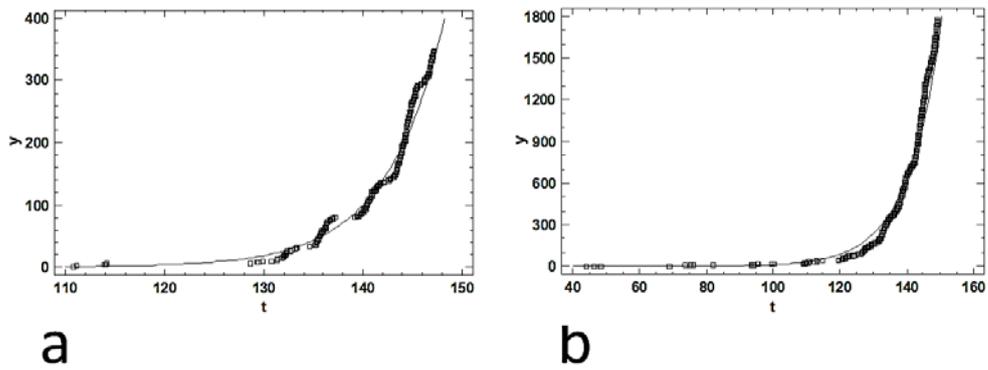

Figure 10. Number of bacteria bearing optimal solution (*y*) as a function of time (*t*) in the 0/1 knapsack optimization problema (12): (a) standard BAGA (b) BAGA algorithm with penalty.

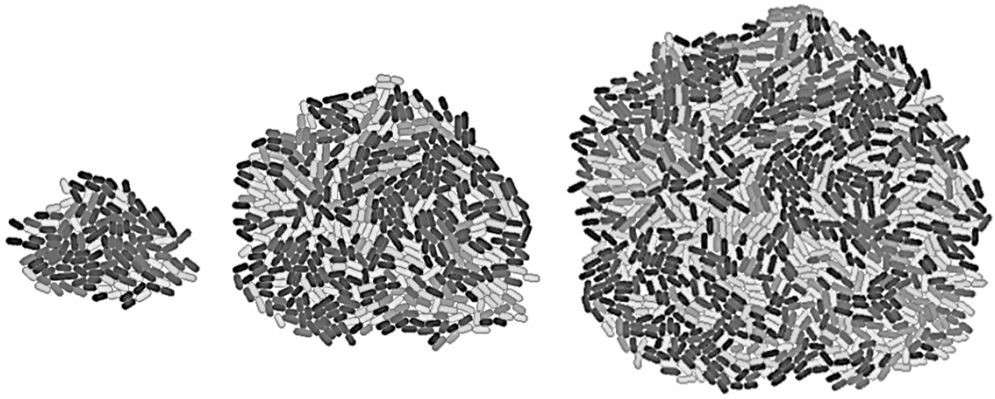

Figure 11. Growth of the colony of synthetic bacteria in the Hamiltonian path problem. The size of the colony is 125 bacteria in $t$=167.71 (left), 503 bacteria in $t$=215.16 (center) and 1017 bacteria in $t$=241.06 (right).

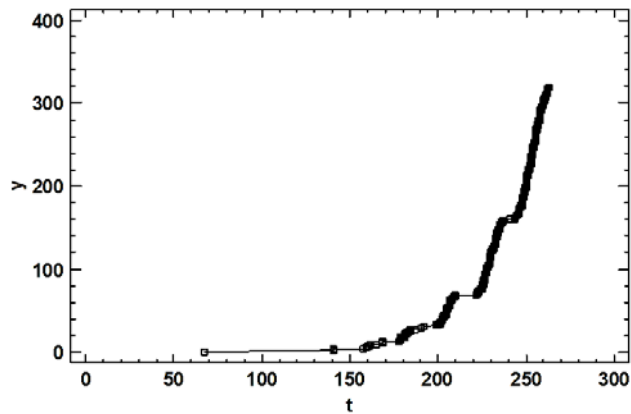

Figure 12. Number of bacteria bearing optimal solution ($y$) as a function of time ($t$) in the Hamiltonian path problem with three nodes. In the experiment, the regression equation of the growth model has a function $y = e^{-3.328+0.034t}$ ($p$-value=0.0000).

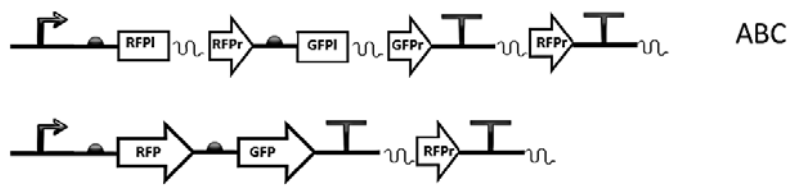

Figure 13. Hamiltonian path solution.